\documentclass[10pt,twocolumn,letterpaper]{article}

\usepackage{cvpr}
\usepackage{times}
\usepackage{epsfig}
\usepackage{graphicx}
\usepackage{amsmath}
\usepackage{amssymb}

\usepackage{amsfonts}
\usepackage{subfigure}
\usepackage{float}
\usepackage{multirow}
\usepackage{tabularx}
\usepackage{makecell}
\usepackage{dsfont}
\usepackage{threeparttable}
\usepackage{color}
\usepackage[justification=centering]{caption}
\usepackage{epstopdf}
\usepackage[lined,boxed,commentsnumbered, ruled]{algorithm2e}
\usepackage[normalem]{ulem}
\usepackage{verbatim}
\graphicspath{{./figs/}}

\newcommand{\citep}{\cite}

\definecolor{blue-violet}{rgb}{0.54, 0.17, 0.89}
\definecolor{mygreen}{rgb}{0.0, 0.5, 0.0}
\definecolor{awesome}{rgb}{1.0, 0.13, 0.32}
\definecolor{bostonuniversityred}{rgb}{0.8, 0.0, 0.0}

\newcounter{ToDo}
\newcounter{guocomm}

\usepackage{soul}
\definecolor{darkblue}{rgb}{0,0,0.5}
\setulcolor{blue-violet}

\cvprfinalcopy
\def\cvprPaperID{} 

\begin{document}
\title{ Kernel Sparse Subspace Clustering on Symmetric Positive Definite
Manifolds}
\author{Ming Yin$^1$, Yi Guo$^2$, Junbin Gao$^3$, Zhaoshui He$^1$ and Shengli Xie$^1$\\
{\small $^1$School of Automation, Guangdong University of Technology, Guangzhou 510006, P. R. China}\\
{\small $^2$Commonwealth Scientific and Industrial Research Organisation, North Ryde, NSW 1670, Australia}\\
{\small $^3$ 
The University of Sydney Business School, The University of Sydney, Camperdown, NSW 2006, Australia}\\
{\tt\small \{yiming,zhshhe,shlxie\}@gdut.edu.cn; yi.guo@csiro.au; junbin.gao@sydney.edu.au}
}
\maketitle
\begin{abstract}
  Sparse subspace clustering (SSC), as one of the most successful subspace clustering methods, has achieved notable clustering accuracy in computer vision tasks. However, SSC applies only to vector data in Euclidean space. As such, there is still no satisfactory approach to solve  subspace clustering by {\it self-expressive} principle for symmetric positive definite (SPD) matrices which is very useful in computer vision. In this paper, by embedding the SPD matrices into a Reproducing Kernel Hilbert Space (RKHS), a kernel subspace clustering method is constructed on the SPD manifold through an appropriate Log-Euclidean kernel, termed as kernel sparse subspace clustering on the SPD Riemannian manifold  (KSSCR). By exploiting the intrinsic Riemannian geometry within data, KSSCR can effectively characterize the geodesic distance between SPD matrices to uncover the underlying subspace structure. Experimental results on two famous database demonstrate that the proposed
method achieves better clustering results than the state-of-the-art approaches.
\end{abstract}

\section{Introduction}
Despite the majority of subspace clustering methods \cite{Vidal2011} show good performance in various applications, the similarity among data points is measured in the original data domain. Specifically, this similarity is often measured between the sparse \citep{DonohoEladTemlyakov2006} or low-rank representations \cite{LiuLinYanSunMa2013} of  data points , by exploiting the {\it self-expressive} property of the data, in terms of Euclidean alike distance. In general, the representation for each data point is its linear regression coefficient on the rest of the data subject to sparsity or low-rank constraint.  Unfortunately, this assumption may not be always true for many high-dimensional data in real world where data may be better modeled by nonlinear manifolds \citep{HarandiHartleyLovellSanderson2014,JayasumanaHartleySalzmannLiHarandi2013,NguyenYangShenSun2015}. In this case, the {\it self-expressive} based algorithms, such as Sparse Subspace Clustering (SSC) \cite{ElhamifarVidal2013} and Low Rank Subspace Clustering (LRSC) \cite{VidalFavaro2014}, are no longer applicable. For example, the human facial images are regarded as samples from a nonlinear submanifold \citep{LiWangZuoZhang2013,PatelVidal2014}.

Recently, a useful image and video descriptor, the covariance descriptor which is a symmetric positive definite (SPD) matrix\citep{TuzelPorikliMeer2006},  has attracted a lot of attention. By using this descriptor, a promising classification performance can be achieved \citep{HarandiHartleyLovellSanderson2014}. However, the traditional subspace learning mainly focuses on the problem associated with vector-valued data. It is known that SPD  matrices form a Lie group, a well structured Riemannian manifold. The naive way of vectorizing SPD matrices first and applying any of the available vector-based techniques usually makes the task less intuitive and short of proper interpretation. The underlying reason is the lack of vector space structures in Riemannian manifold. That is, the direct application of linear reconstruction model for this type of data will result in inaccurate representation and hence compromised performance.

To handle this special case, a few solutions have been recently proposed to address sparse coding problems on Riemannian manifolds, such as \citep{CherianSra2014,HoXieVemuri2013,SivalingamBoleyMorellasPapanikolopoulos2014}. While for subspace clustering, a nonlinear LRR model is proposed to extend the traditional LRR from Euclidean space to Stiefel manifold \cite{YinGaoGuo2015}, SPD manifold \cite{FuGaoHongTien2015} and abstract Grassmann manifold \citep{WangHuGaoSunYin2015} respectively. The common strategy used in the above research is to make the data self-expressive principle work by using the logarithm mapping ``projecting'' data onto the tangent space at each data point where a ``normal'' Euclidean linear reconstruction of a given sample is well defined \citep{TuzelPorikliMeer2008}. This idea was first explored by Ho \emph{et al}. \citep{HoXieVemuri2013} and they  proposed a nonlinear generalization of sparse coding to handle the non-linearity of Riemannian manifolds, by flattening the manifold using a fixed tangent space.

For the special Riemannian manifold of SPD, many researchers \citep{CherianSra2014,SivalingamBoleyMorellasPapanikolopoulos2014} took advantage of a nice property of this manifold, namely  that the manifold is closed under positive linear combination, and exploited appropriate nonlinear metrics such as log-determinant divergence to measure errors in the sparse model formulation. This type of strategies fall in the second category of methods dealing with problems on manifolds by embedding manifold onto a larger flatten Euclidean space. For example, the LRR model on abstract Grassmann manifold is proposed based on the embedding technique \citep{WangHuGaoSunYin2014} and its kernelization \citep{WangHuGaoSunYin2015a}.

\begin{figure}[ht]
\centering
 \includegraphics[width=0.4 \textwidth]{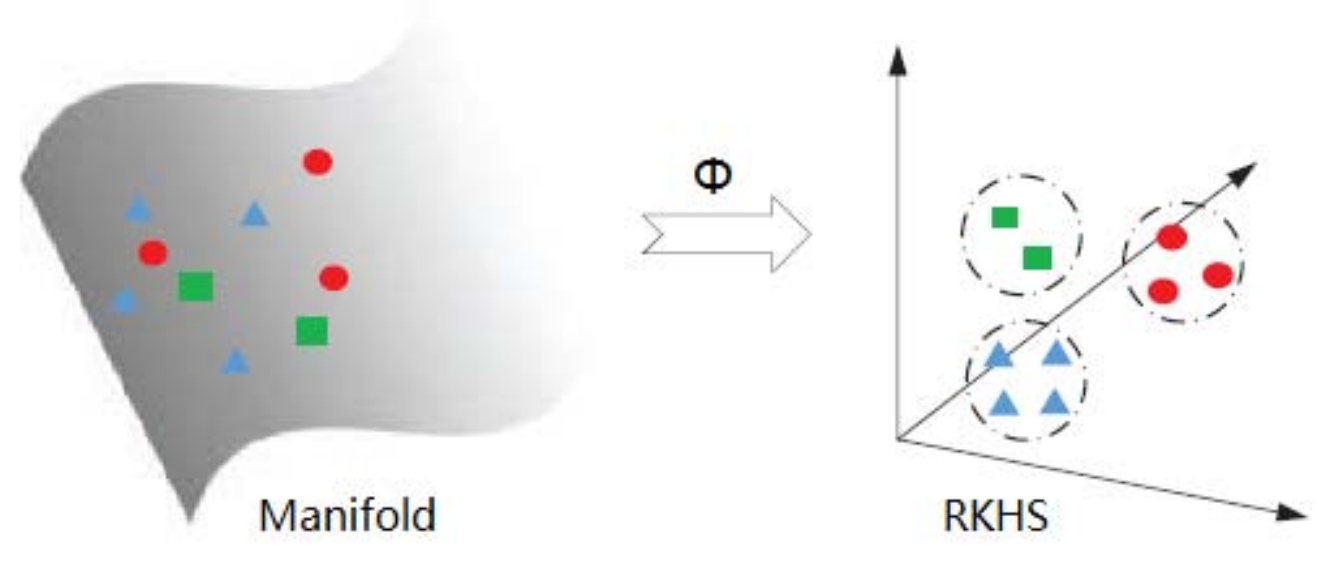}
\caption{Overview of the proposed kernel sparse subspace clustering method. Using the SPD kernel, data is mapped onto a high-dimensional Reproducing Kernel Hilbert Space (RKHS) so as to attain better representations for clustering. }\label{fig1}
\end{figure}

It is noteworthy that a nonlinear extension of SSC method for manifold clustering has been proposed in \cite{PatelVidal2014}. Unfortunately, the authors simply used the kernel trick to map data onto a high-dimensional feature space for vector data, not considering the intrinsic geometric structure. To rectify this, in this paper, we propose to use a kernel feature mapping to embed the SPD Riemannian manifold into a high dimension feature space and preserve its intrinsic Riemannian geometry within data. We call this method kernel sparse subspace clustering on Riemannian manifold (KSSCR). An overview of the proposed method is illustrated in Fig. \ref{fig1}. Different from the work in \cite{PatelVidal2014}, our motivation is to map SPD matrices into Reproducing Kernel Hilbert Space (RKHS) with Log-Euclidean Gaussian kernel based on Riemannian metric. As a result, the linear reconstruction can be naturally implemented in the feature space associated with the kernel where the original SSC can be applied. The proposed method not only effectively characterizes the geodesic distance between pair of SPD matrices but also uncovers the underlying low-dimensional subspace structure.

The remainder of the paper is organized as follows. In Section \ref{SectionII}, we give a brief review on related work. Section \ref{SectionIII} is dedicated to introducing the novel kernel sparse subspace clustering on Riemannian manifold. The experimental results are given in Section \ref{SectionIV} and Section \ref{SectionV} concludes the paper with a brief summary.

\section{Related Work}\label{SectionII}
Before introducing the proposed model, in this section, we briefly review the recent development of subspace clustering methods \citep{ElhamifarVidal2013} and the analysis of Riemannian geometry of SPD Manifold \citep{PennecFillardAyache2006}. Throughout the paper,  capital letters denote matrices (e.g., $X$) and bold lower-case letters denote column vectors (e.g., $\textbf{x}$). ${x}_i$ is the $i$-th element of vector $\textbf{x}$. Similarly, $X_{ij}$ denotes the $(i,j)$-th entry of matrix $X$. $\| \textbf{x} \|_1 = \sum\limits_{i} |{x}_i|$ and $\|\textbf{x} \|_2 = \textbf{x}^T\textbf{x}$  are the $\ell_1$ and $\ell_2$ norms respectively, where $^T$ is the transpose operation. $ \left\| \cdot \right\|_F $ is the matrix Frobenius norm defined as $\left\| X \right\|_F^2 = \sum\limits_{i } {\sum\limits_{j} {\left| {X_{ij} } \right|^2 }}$.
The space of $d \times d$ SPD matrices is denoted by $\mathcal{S}_d^+$. The tangent space at a point $X$ on $\mathcal{S}_d^+$ is defined by $T_X \mathcal{S}_d^+$, which is a vector space including the tangent vectors of all possible curves passing over $X$.

\subsection{Subspace Sparse Representation}
Sparse representation, which has been proved to be a useful tool for representing and compressing high-dimensional signals, provides a statistical model for finding efficient and compact signal representations. Among them, SSC \citep{ElhamifarVidal2013} is of particular interests as it clusters data points to a union of low-dimensional subspaces without referring to a library. Let $X= [\mathbf{x}_1, \mathbf{x}_2, ..., \mathbf{x}_N] \in \mathbb{R}^{d \times N}$ be a matrix of data. Each datum $\mathbf x_i$ is drawn from a low-dimensional subspace denoted by $\mathcal S_j$ for $j\in[1,\ldots,k]$. By exploiting the {\it{self-expressive}} property of the data, the formulation of SSC is written as follows,
\begin{align}
\mathop{\min}\limits_{\textrm{C}} ~\| \textrm{C} \|_1,  ~\textrm{ s.t. }{X} = {X}\textrm{C}+E, ~\textrm{diag(C)} =0. \label{SSC}
\end{align}
where $\textrm{C}= [\mathbf{c}_1, \mathbf{c}_2, ..., \mathbf{c}_N] \in \mathbb{R}^{N \times N}$ is the coefficient matrix whose column $\mathbf{c}_i$ \footnote{Matrix $\mathbf C$ is bold, while other matrices are not.}is the sparse representation vector corresponding to the $i$-th data point. $E$ denotes the reconstruction error components.

As to the coefficient matrix $\mathbf C$, besides the interpretation as new sparse representation of the data, each element $C_{ij}$ in $\mathbf C$ can also be regarded as a similarity measure between the data pair $\mathbf x_i$ and $\mathbf x_j$. In this sense, $\textrm{C}$ is sometimes called an affinity matrix. Therefore, a clustering algorithm such as K-means can be subsequently applied to $\textrm{C}$ for the final segmentation solution. This is a common practice of subspace clustering based on finding new representation.

\subsection{Spare Representation on SPD matrices}
Since SPD matrices belong to a Lie group which is a Riemannian manifold \citep{ArsignyFillardPennecAyache2007}, it cripples many methods that rely on linear reconstruction. Generally, there are two methods to deal with the non-linearity of Riemannian manifolds. One is to locally flatten the manifold to tangent spaces\citep{TuzelPorikliMeer2008}. The underlying idea is to exploit the geometry of the manifold directly. The other is to map the data into a feature space usually a Hilbert space \citep{JayasumanaHartleySalzmannLiHarandi2013}. Precisely, it is to project the data into RKHS through kernel mapping \citep{HarandiSandersonHartleyLovell2012}. Both of these methods are seeking a transformation so that the linearity re-emerges.

A typical example of the former method is the one in \citep{HoXieVemuri2013}. Let $X$ be a SPD matrix and hence a point on $\mathcal{S}_d^+$. $\mathbb{D} = \{D_1, D_2, ..., D_N \}, D_i \in \mathcal{S}_d^+$ is a dictionary. An optimization problem for sparse coding of  $X$ on a manifold $\mathcal{M}$ is formulated as follows
\begin{align}
\mathop{\min}\limits_{\mathbf{w}} ~\lambda\|\mathbf{w}\|_1 +  \left\| \sum\limits_{i = 1}^N {w}_{i}\mathbf{log}_{X}(D_i)  \right\|_{X}^2, ~\textrm{ s.t. } \sum\limits_{i = 1}^N w_{i}=1,\label{SCManifold}
\end{align}
where $\mathbf{log}_X(\cdot)$ denotes Log map from SPD manifold to a tangent space at $X$,  $\mathbf{w} = [w_1, w_2,..., w_N]$ is the sparse vector and $\|\cdot\|_X$ is the norm associated with $T_X \mathcal{S}_d^+$. Because $\mathbf{log}_X(X)=\textbf{0}$, the second term in Eq.\eqref{SCManifold} is essentially the error of linearly reconstructing $\mathbf{log}_X(X)$ by others on the tangent space of $X$, As this tangent space is a vector space, this reconstruction is well defined. As a result, the traditional sparse representation model can be performed on Riemannian manifold.

However, it turns out that quantifying the reconstruction error is not at all straightforward. Although $\ell_2$-norm is commonly used in the Euclidean space, using  Riemannian metrics would be better in $\mathcal{S}_d^+$ since they can accurately measure the intrinsic distance between SPD matrices. In fact, a natural way to measure closeness of data on a Riemannian manifold is geodesics, i.e.  curves analogous to straight lines in $\mathds{R}^n$. For any two data points on a manifold, geodesic distance is the length of the shortest curve on the manifold connecting them. For this reason, the affine invariant Riemannian metric (AIRM) is probably the most popular Riemannian metric defined as follows \citep{PennecFillardAyache2006}. Given $X \in \mathcal{S}_d^+$, the AIRM of two tangent vectors $\mathbf{v}, \mathbf{w} \in T_X \mathcal{S}_d^+$ is defined as
\begin{align*}
&\langle \mathbf{v}, \mathbf{w}  \rangle = \langle X^{-1/2}\mathbf{v} X^{-1/2}, X^{-1/2} \mathbf{w} X^{-1/2} \rangle \\
&= \textrm{tr} ( X^{-1} \mathbf{v}  X^{-1} \mathbf{w} ).
\end{align*}
The geodesic distance between points $X, Y \in \mathcal{S}_d^+$  induced from AIRM is then
\begin{align}
&\delta_g (X, Y) = \|\mathbf{log} (X^{-1/2}Y X^{-1/2}) \|_F . \label{gAIRM}
\end{align}

\section{Kernel Subspace Clustering on SPD Matrices}\label{SectionIII}
Motivated by the above issues, in this section, we propose a novel kernel sparse subspace clustering algorithm which enables SSC to  handle data on Riemannian manifold by incorporating the intrinsic geometry of the manifold. The idea is quite simple but effective: map the data points into RKHS first and then perform SSC with some modifications in RKHS. Compared with the original SSC, the advantages of this approach include simpler solutions and better representation due to the capability of learning the underlying nonlinear structures. The following is the detail of our method. Given a data set $\mathcal{X}= [X_1,X_2,..., X_N]$ on SPD manifold, we seek its sparse representation via exploiting the {\it self-expressive} property of the data. Thus, the objective of our kernel sparse subspace representation algorithm on Riemannian manifold is formulated as follows
\begin{align}
&\mathop{\min}\limits_{\textrm{C}} ~\lambda\|\textrm{C}\|_1 + \sum\limits_{i = 1}^N \left\|\phi(X_i)- \sum\limits_{j = 1}^N c_{ij} \phi(X_j) \right\|_F^2, \notag \\ &~\textrm{{s.t.}~diag(\textrm{C})} =0. \label{KSSCR}
\end{align}
where $\phi( \cdot )$ denotes a feature mapping function that projects SPD matrices into RKHS such that $\langle \phi(X),\phi(Y) \rangle = \kappa(X,Y)$ where $\kappa(X,Y)$ is a positive definite (PD) kernel.

\subsection{Log-Euclidean Kernels for SPD Matrices}
Although locally flattening Riemannian manifolds via tangent spaces \citep{HoXieVemuri2013} can  handle their non-linearity, it inevitably  leads to very demanding computation due to switching back and forth between tangent spaces and the manifold. Furthermore, linear reconstruction of SPD matrices  is not as natural as in Euclidean space and this may incur errors.

The recent work in \citep{HarandiHartleyLovellSanderson2014} shows the property of Stein divergence is akin to AIRM. Furthermore, a PD kernel can be induced from Stein divergence under some conditions \citep{Sra2012}. Concretely, a Stein metric\citep{Sra2012},  also known as Jensen-Bregman LogDet divergence (JBLD) \citep{CherianSraBanerjeePapanikolopoulos2013},  derived from Bregman matrix divergence is given  by,
\begin{align*}
&\delta_s (X, Y) = \mathbf{log}|\frac{X+ Y}{2}|- \frac1{2}\mathbf{log}|XY|,
\end{align*}
where $|\cdot |$ denotes determinant.  Accordingly a kernel function based on  Stein divergence  for SPD $\mathcal{S}_d^+$ can be defined as
$ \kappa_s(X,Y) = \textrm{exp}\{-\beta \delta_s (X, Y)  \}$, though it is guaranteed to be positive definite only when $\beta \in \{ \frac1{2},1,...,\frac{d-1}{2}\}$ or $\beta > \frac{d-1}{2}$ \citep{Sra2012}.

However, the problem is that Stein divergence is only an approximation to Riemannian metric and cannot be a PD kernel without more restricted conditions \citep{LiWangZuoZhang2013}. If one uses this kernel, the reconstruction error for Riemannian metric will be incurred. To address this problem, a family of Log-Euclidean kernels were proposed in \citep{LiWangZuoZhang2013}, i.e., a polynomial kernel, an exponential kernel and a Gaussian kernel, tailored to model data
geometry more accurately. These Log-Euclidean kernels were proven to well characterize the true geodesic distance between SPD matrices, especially the Log-Euclidean Gaussian kernel
\[ \kappa_g(X,Y) = \textrm{exp}\{-\gamma \|\mathbf{log}(X)- \mathbf{log}(Y)\|_F^2  \},
\]
which is a PD kernel for any  $\gamma > 0$. Owing to its superiority, we select Log-Euclidean Gaussian kernel to transform the SPD matrices into RKHS.

\subsection{Optimization}\label{SectionIII2}
{In this subsection, we solve the objective of kernel sparse subspace learning in \eqref{KSSCR} via alternating direction method of multipliers (ADMM) \citep{BoydParikhChuPeleatoEckstein2011}. Expanding the Frobenius norm in \eqref{KSSCR} and applying kernel trick leads to the following equivalent problem
\begin{align}
&\mathop{\min}\limits_{\textrm{C}} ~\lambda\|\textrm{C}\|_1- 2 \textrm{tr}(K\textrm{C}) +\textrm{tr}(\textrm{C}K\textrm{C}^T), \\ \notag &\quad\textrm{s.t.~diag}(\textrm{C}) =0,
\end{align}
where $K= \{k_{ij}\}_{i,j = 1}^n$ is the kernel Gram matrix, and $k_{ij}=  \kappa_g(X_j,X_i)=\phi(X_j)^T \phi(X_i)$.

By introducing an auxiliary matrix $A$, the above problem can be rewritten as follows
\begin{align}
&\mathop{\min}\limits_{\textrm{C},A} ~\lambda\|\textrm{C}\|_1- 2 \textrm{tr}(KA) +\textrm{tr}(AKA^T), \notag\\ &\quad\textrm{s.t.} ~ A = \textrm{C}- \textrm{diag}(\textrm{C}). \label{solve_Kscc}
\end{align}

The augmented Lagrangian is then
\begin{align}
\mathcal{L}(\textrm{C},A)= \mathop{\min}\limits_{\textrm{C},A}  ~\lambda\|\textrm{C}\|_1- 2 \textrm{tr}(KA) +\textrm{tr}(AKA^T) \notag\\
+ \frac{\rho}{2} \| A- \textrm{C} + \textrm{diag}(\textrm{C})\|_F^2 +\textrm{tr}(\Delta^T( A- \textrm{C} + \textrm{diag}(\textrm{C}))), \label{7}
\end{align}
on which ADMM is carried out. Note that $\Delta$ is a Lagrangian multiplier matrix with compatible dimension.

The optimization algorithm is detailed as follows.
\begin{enumerate}
\item Update $A$.
\begin{align}
&\mathop{\min}\limits_{A} - 2 \textrm{tr}(KA)+ \frac{\rho}{2} \| A- \textrm{C}+ \textrm{diag}(\textrm{C})\|_F^2 \notag\\
& + \textrm{tr}(AKA^T)+\textrm{tr}(\Delta^T( A- \textrm{C} + \textrm{diag}(\textrm{C})))\label{sloverA1}
\end{align}
Let $\widetilde{\textrm{C}}=\textrm{C} - \textrm{diag}(\textrm{C})$, the subproblem can be formulated by,
\begin{align}
&\mathop{\min}\limits_{A} - 2 \textrm{tr}(KA) +\textrm{tr}(AKA^T) \notag\\
&+ \frac{\rho}{2} \| A- \widetilde{\textrm{C}} \|_F^2 +\textrm{tr}(\Delta^T( A- \widetilde{\textrm{C}}))\label{sloverA2}
\end{align}

Setting the derivative  w.r.t. $A$ to zero results in a closed-form solution to subproblem \eqref{sloverA1} given by
\begin{align}\label{E:updateA}
A_{t+1} =(2K+\rho \widetilde{\textrm{C}}_t-\Delta_t) (2K+\rho\mathbf{I})^{-1}.
\end{align}
where $\mathbf{I}$ is an identity matrix.

\item Update $\textrm{C}$.
\begin{align}
&\mathop{\min}\limits_{\textrm{C}}  ~\lambda\|\textrm{C}\|_1+ \frac{\rho}{2} \| A- \textrm{C}+ \textrm{diag}(\textrm{C})\|_F^2 \notag\\
&+\textrm{tr}(\Delta^T( A- \textrm{C} + \textrm{diag}(\textrm{C}))).
\end{align}
The above subproblem has the following closed-form solution given by shrinkage operator
\begin{align}
\textrm{C}_{t+1} = \mathbf{J}- \textrm{diag}(\mathbf{J}), \\ \mathbf{J}=\mathcal{S}_{\frac{\lambda}{\rho}}(A_{t+1}+\frac{\Delta_t}{\rho}).
\end{align}
where $\mathcal{S}_{\eta}(\cdot)$ is a shrinkage operator acting on each element of the given matrix, and is defined as $\mathcal{S}_{\eta}(v) = \textrm{sgn}(v)\textrm{max}(|v|- \eta, 0) $.

\item Update $\Delta$.
\begin{align}
\Delta_{t+1} = \Delta_t + \rho ( A_{t+1}- \textrm{C}_{t+1} + \textrm{diag}(\textrm{C}_{t+1})) .
\end{align}
\end{enumerate}

These steps are repeated until $\|A_t- \textrm{C}_t \|_{\infty} \leq \epsilon, \|A_{t+1}- A_{t}\|_{\infty} \leq \epsilon $.

\subsection{Subspace Clustering}
As discussed earlier, $\textrm{C}$ is actually a new representation of data learned found by using data {\it{self-expressive}} property. After solving problem \eqref{KSSCR}, the next step is to segment $\textrm{C}$ to find the final subspace clusters. Here we apply a spectral clustering method to the affinity matrix given by $(|\textrm{C}|+|\textrm{C}^T|)/2$ to separate the data into clusters, which is equivalent to subspaces. The complete clustering method is outlined in Algorithm \ref{Alg1}.
\begin{algorithm}
\caption{Kernel Subspace Clustering on SPD Matrices }
\SetKwData{Index}{Index}
\KwIn{ $\mathcal{X}= [X_1,X_2,..., X_N]$ , $\gamma$, $\lambda$ and $\rho$.}
\BlankLine
\textbf{Steps:}
\begin{enumerate}
\item Solve \eqref{KSSCR} by ADMM explained in Section \ref{SectionIII2},
and obtain the optimal solution $ \textrm{C}^* $.
\item Compute the affinity matrix $\mathcal{W}$ by,
\begin{align*}
 \mathcal{W} = \left(|\textrm{C}^*|+ |\textrm{C}^*|^T\right)/2.
\end{align*}
\item Apply normalized spectral clustering method to $\mathcal{W}$  to obtain the final clustering solution $\mathcal C$.
\end{enumerate}
\KwOut{the clustering solution {$\mathcal C$}.}
\label{Alg1}
\end{algorithm}

\subsection{Complexity Analysis and Convergence}
The computational cost of the proposed algorithm is mainly determined by the steps in ADMM. The total complexity of KSSCR is, as a function of the number of data points, $\mathcal{O}(\frac13N^3+t N^2)$ where $t$ is the total number of iterations. The soft thresholding to update the sparse matrix $\textrm{C}$ in each step is relatively cheaper, much less than $\mathcal{O}(N^2)$.  For updating $A$ we can pre-compute the Cholesky decomposition of $(2K+\rho\mathbf I)^{-1}$ at the cost of less than $\mathcal{O}(\frac12 N^3)$, then compute new $A$ by using \eqref{E:updateA} which has a complexity of $\mathcal{O}(N^2)$ in general.

The above proposed ADMM iterative procedure to the augmented Lagrangian problem \eqref{7} satisfies the general condition for the convergence theorem in \citep{LinLiuLi2015}.

\section{Experimental Results} \label{SectionIV}
In this section, we present several experimental results to demonstrate the  effectiveness of KSSCR. To comprehensively evaluate the performance of KSSCR, we tested it on texture images and  human faces.  Some sample images of test databases are shown in Figure \ref{fig2}.  The clustering results are shown in Section \ref{sectionIV2} and Section \ref{sectionIV3} respectively.  All test data lie on Riemannian manifold.
\begin{figure}[ht]
\centering
  \subfigure[]{\includegraphics[width=0.23 \textwidth]{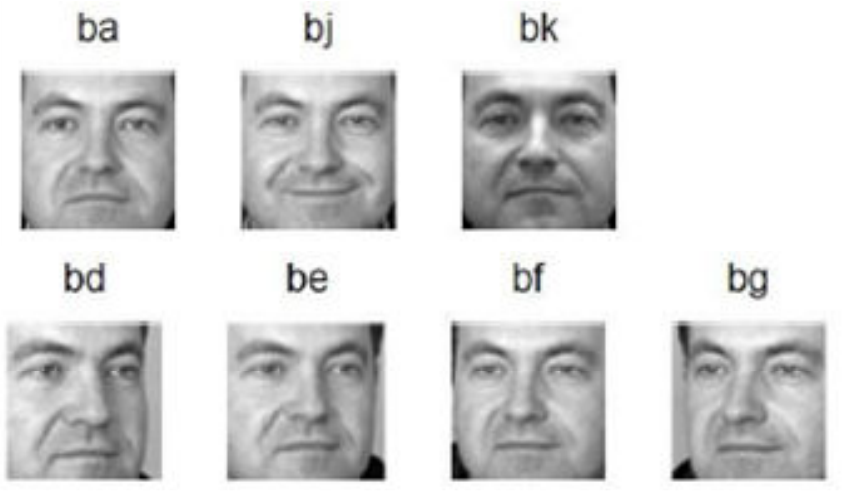}}
   \subfigure[]{\includegraphics[width=0.12 \textwidth]{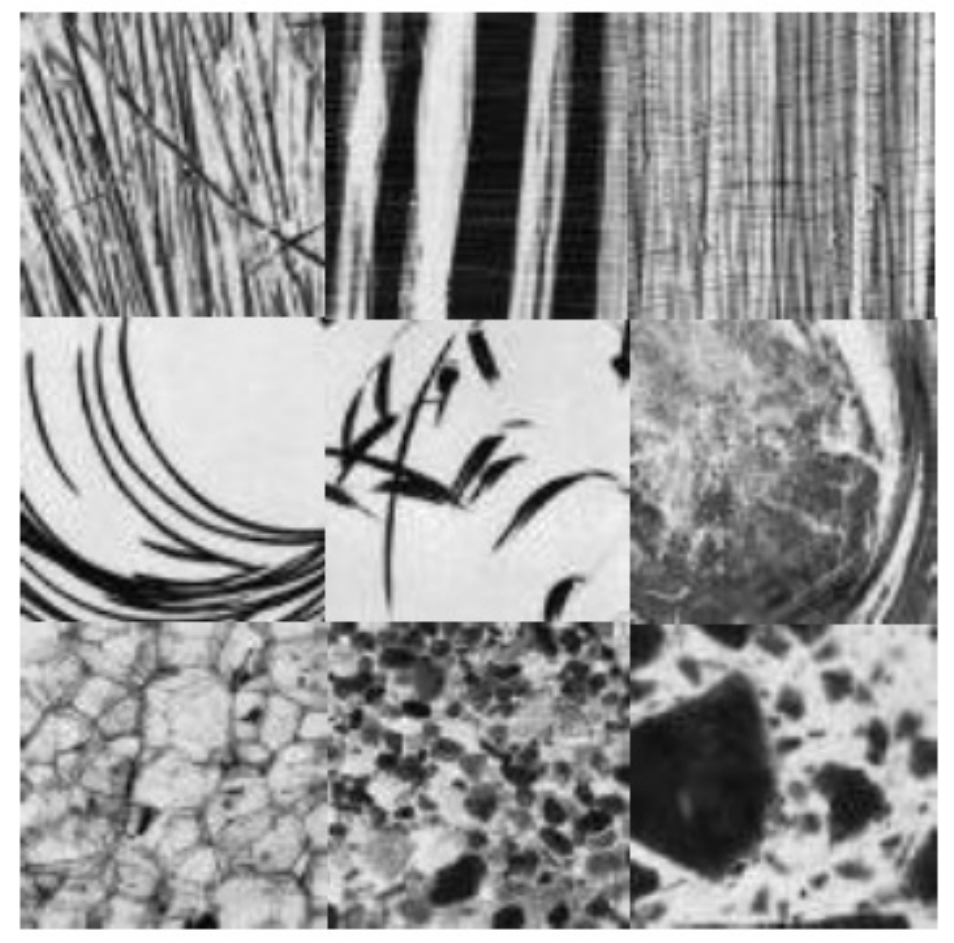}}
   \caption{Samples on the FERET (a) and Brodatz (b) database. }\label{fig2}
\end{figure}

In this work, we adopt two criteria, i.e. Normalized Mutual Information (NMI) and subspace clustering accuracy, to quantify the clustering performance more precisely. NMI aims to measure how similar different cluster sets are. Meanwhile, to extensively assess how the proposed algorithm improves the performance of data clustering, the following four state-of-the-art subspace clustering methods are compared against:
\begin{enumerate}
 \item  Sparse Subspace Clustering(SSC) \citep{ElhamifarVidal2013}.
 \item  Low-rank Representation(LRR) \citep{LiuLinYu2010}.
 \item  Low-Rank Subspace Clustering (LRSC) \citep{VidalFavaro2014}, which aims to seek a low-rank representation by decomposing the corrupted data matrix as the sum of a clean and self-expressive dictionary.
 \item Kernel SSC on Euclidean space (KSSCE) \cite{PatelVidal2014}, which embeds data onto to a nonlinear manifold by using the kernel trick and then apply SSC based on Euclidean metric.
\end{enumerate}
We also included  K-means clustering algorithm (K-means) as a baseline.

\subsection{Texture Clustering}\label{sectionIV2}
In this subsection, a subset of the Brodatz database \citep{LiWangZuoZhang2013}, i.e., 16-texture (`16c') mosaic, was chosen for clustering performance evaluation. There are 16 objects in this subset in which each class contains only one image. Before clustering, we downsampled each image to 256 $\times$ 256 and then split into 64 regions of size 32 $\times$ 32. To obtain their region covariance matrices (RCM), a feature vector $f(x,y)$ for any pixel $I(x,y)$ was extracted, e.g., $f(x,y) = (I(x,y), |\frac{\partial I}{\partial x}|, |\frac{\partial I}{\partial y}|, |\frac{\partial^2 I}{\partial x^2}|, |\frac{\partial^2 I}{\partial y^2}|)$. Then, each region can be characterized by a 5 $\times$ 5 covariance descriptor. Totally, 1024 RCM were collected. We randomly chose some data from Brodatz database with the number of clusters, i.e., $N_c$, ranging from 2 to 16. The final performance scores were computed by averaging the scores from 20 trials. The detailed clustering results are summarized in Tables \ref{Tab1} and \ref{Tab2}. We set the parameters as $\lambda = 0.04$ and $\gamma = 0.5$ for KSSCR. Also, the tuned parameters are reported for the results achieved by other methods. The bold numbers highlight the best results.

From the tables, we observe that the proposed method outperforms other methods in most cases while KSSCE achieved the second  best performance. This is due to the nonlinear subspace clustering in Euclidean space without using Riemannian metric. In addition, we find SSC type methods can better discover the intrinsic structure than LRR type ones. In order to show the underlying low-dimensional structure within data, we provide a visual comparison of affinity matrices obtained by different methods in Figure \ref{affMatrix}. Due to the fact that LRSC cannot recover the subspace effectively, we exclude its affinity matrix. It is clear that the affinity matrix achieved by KSSCR effectively reflects the structure of data so as to benefit the subsequent clustering task.
\begin{figure}[ht]
\centering
  \subfigure[]{\includegraphics[width=0.2 \textwidth]{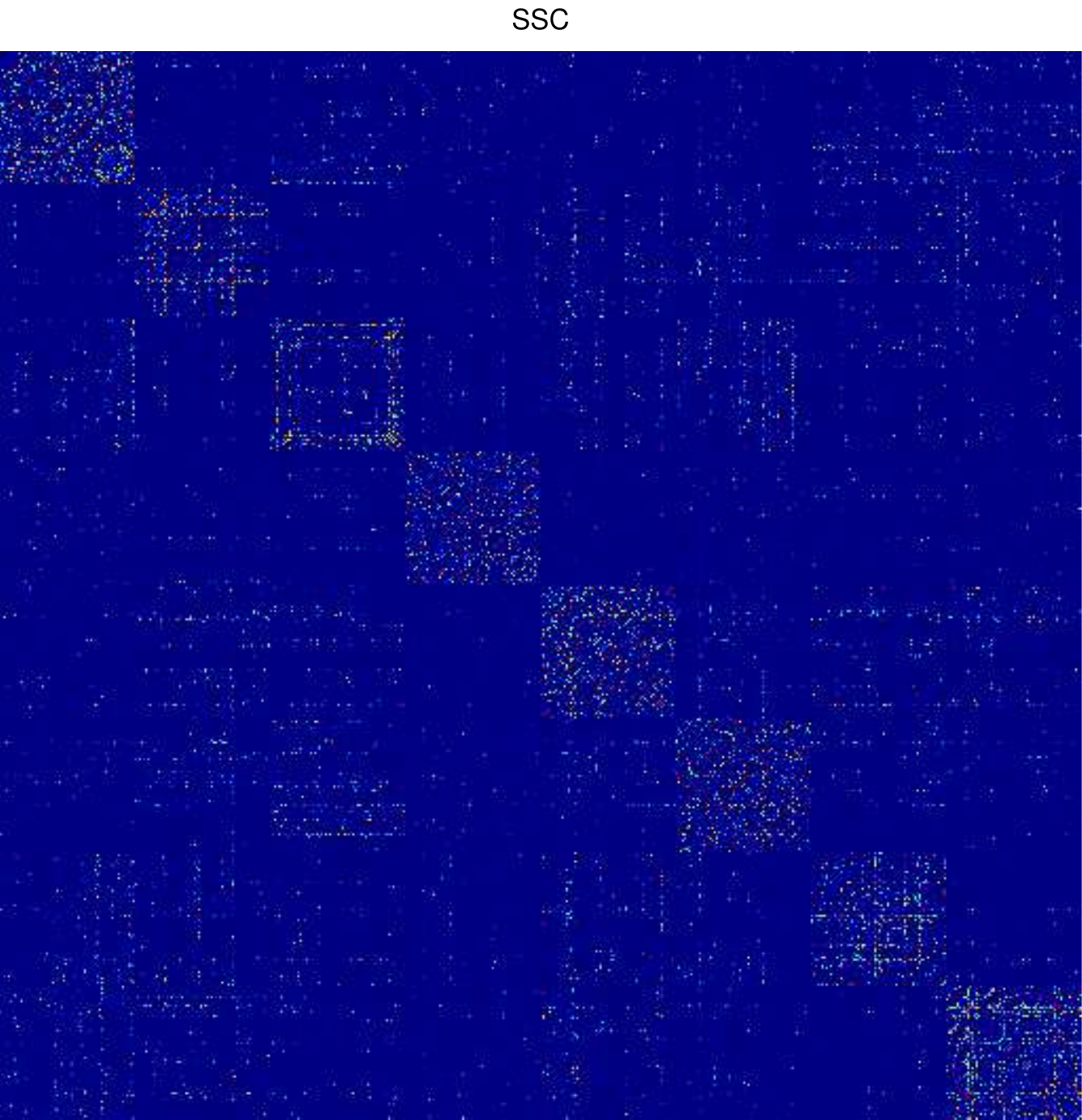}}
  \subfigure[]{\includegraphics[width=0.2 \textwidth]{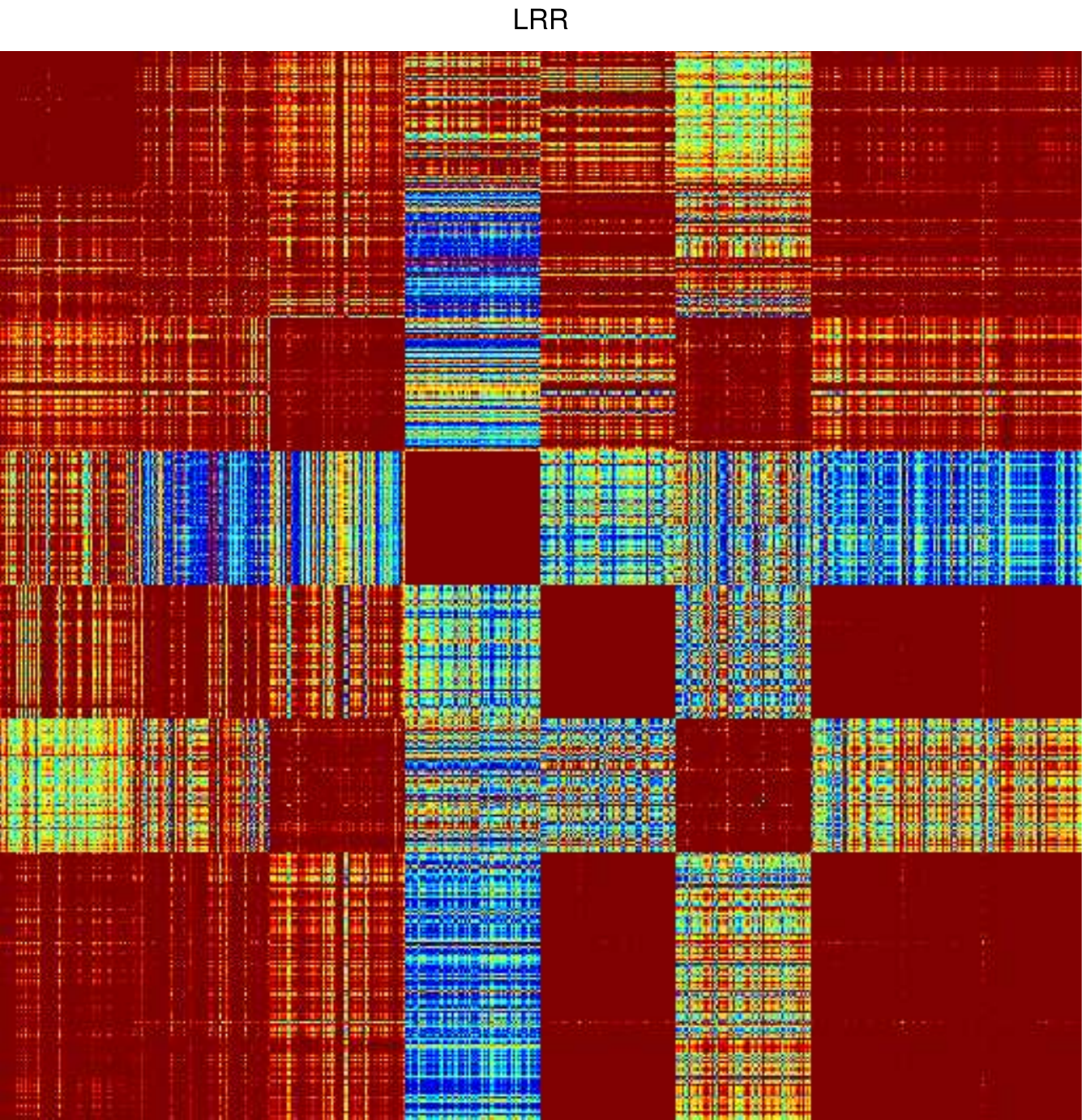}}
  \subfigure[]{\includegraphics[width=0.2 \textwidth]{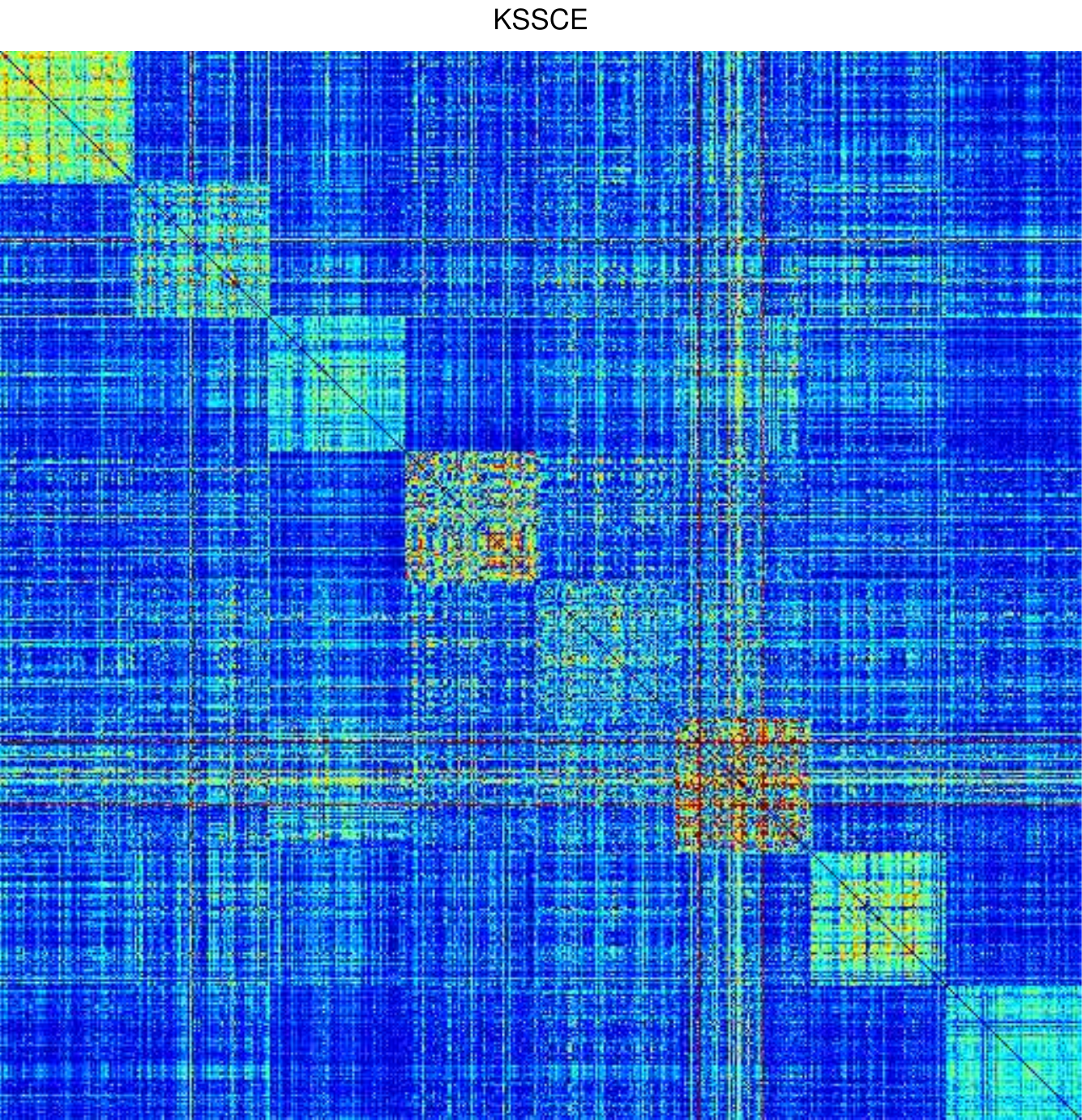}}
  \subfigure[]{\includegraphics[width=0.2 \textwidth]{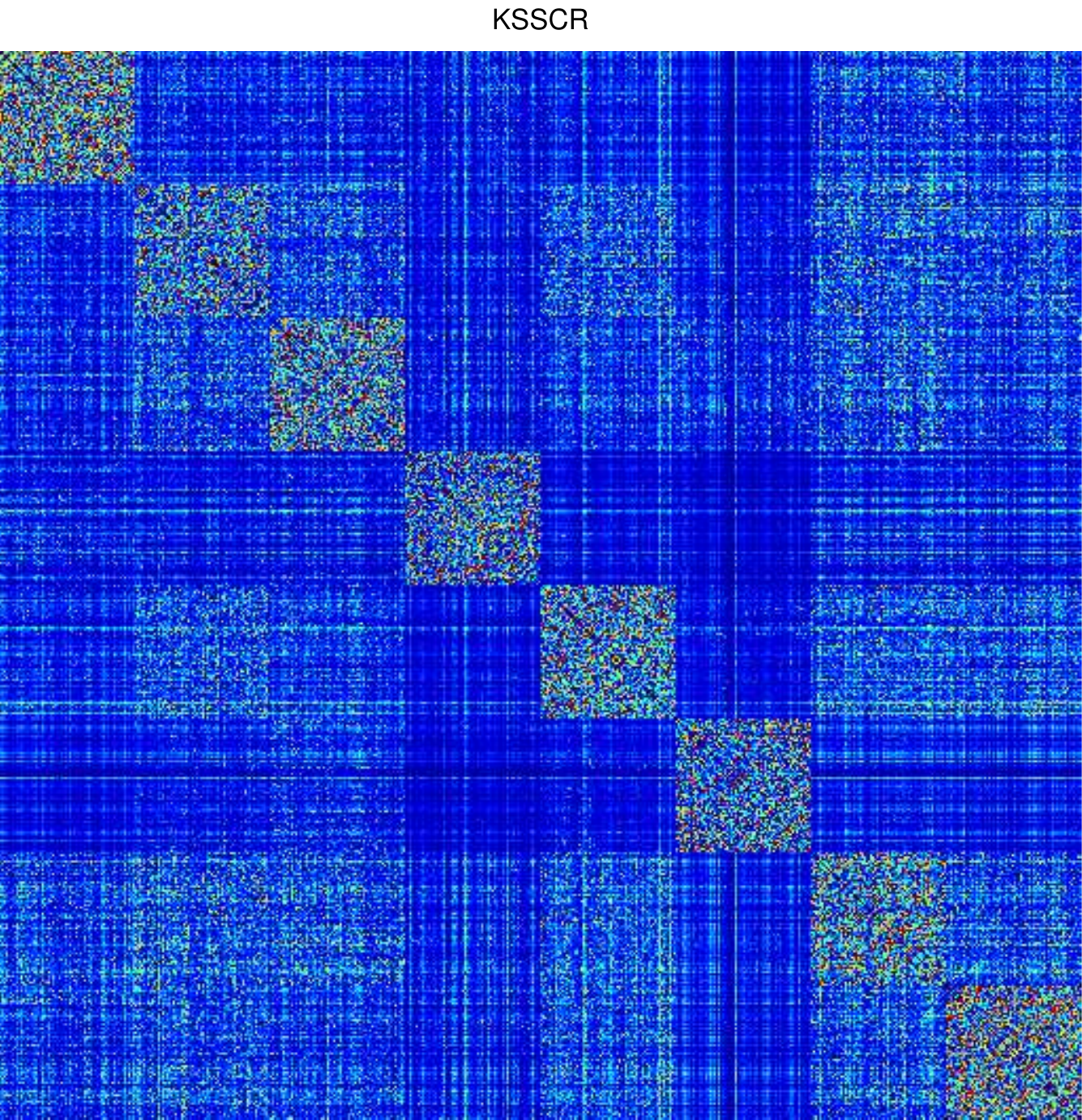}}
   \caption{Examples of affinity matrices on 8-cluster data from the Brodatz database (a) SSC, (b) LRR, (c) KSSCE and (d) KSSCR. }\label{affMatrix}
\end{figure}

\begin{table*}
\begin{center}
\begin{tabular}{|l|c|c|c|c|c|c|}
\hline
$N_c$ & 2 & 4 & 8 & 10 & 12 & 16 \\
\hline\hline
K-means &73.44$\pm$0  &63.28$\pm$ 0  &76.95$\pm$ 0 &61.88$\pm$ 0 &59.24$\pm$ 0 &60.45$\pm$0\\
SSC (100)  &100$\pm$0  &79.30$\pm$0  &78.03$\pm$5.18 &61.77$\pm$0.86    &68.27$\pm$3.26  &60.27$\pm$1.15 \\
LRSC(1, 0.01) &53.90$\pm$0   &$33.98\pm$0  &22.63$\pm$0.30 &21.88$\pm$0.50 &16.87$\pm$ 0.33 &15.95$\pm$0.46\\
LRR(0.4)  &71.88$\pm$0 &42.58$\pm$0  &52.17$\pm$1.66 &53.52$\pm$0.22 &50.77$\pm$0.81 &48.34$\pm$1.84\\
KSSCE(0.4) &78.91 $\pm$ 0 & 53.52 $\pm$ 0  &77.64 $\pm$ 0.93 &72.36 $\pm$ 0.77 & \textbf{82.81$\pm$0.65} & 64.93 $\pm$ 2.89 \\
KSSCR & \textbf{100$\pm$0} & \textbf{85.94$\pm$0} & \textbf{97.85$\pm$0} & \textbf{74.64$\pm$0.08} & {75.7$\pm$0.04} & \textbf{83.62$\pm$0.14}\\
\hline
\end{tabular}
\end{center}
\caption{Clustering results in terms of accuracy (\%) on Brodatz database.}\label{Tab1}
\end{table*}

\begin{table*}
\begin{center}
\begin{tabular}{|l|c|c|c|c|c|c|}
\hline
$N_c$ & 2 & 4 & 8 & 10 & 12 & 16 \\
\hline\hline
K-means &28.70 $\pm$ 0  &45.49 $\pm$ 0  &75.76 $\pm$ 0 &63.76 $\pm$ 0 &62.27 $\pm$ 0 & 63.88 $\pm$0\\
SSC(100)  &100$\pm$0  &63.84$\pm$0     &80.21$\pm$2.60    &64.17 $\pm$0.30   &70.65$\pm$1.93   &67.03$\pm$0.58       \\
LRSC(1, 0.01) & 0.44 $\pm$ 0 &2.08$\pm$0  & 10.86$\pm$0.4    & 14.17 $\pm$0.26 & 10.23$\pm$ 0.43 & 15.37$\pm$ 0.43   \\
LRR(0.4)  &21.24$\pm$0 &15.15 $\pm$ 0  &50.58 $\pm$ 1.65 &48.32 $\pm$0.36 &54.21$\pm$ 0.62 &55.14 $\pm$ 0.89\\
KSSCE(0.4) &28.65 $\pm$ 0 & 35.95 $\pm$ 0  &73.52 $\pm$ 0.53 &70.35 $\pm$ 0.63 & \textbf{79.42$\pm$0.08} &66.35 $\pm$ 2.06 \\
KSSCR & \textbf{100$\pm$0} & \textbf{73.62$\pm$0} & \textbf{95.26$\pm$0} & \textbf{76.71$\pm$0.07} &79.05$\pm$0.04 & \textbf{82.38$\pm$0.11}\\
\hline
\end{tabular}
\end{center}
\caption{Clustering results in terms of NMI (\%) on Brodatz database.}\label{Tab2}
\end{table*}

\subsection{Face Clustering}\label{sectionIV3}
In  this  test, we used the ``b'' subset of FERET database which consists of 1400 images with the size of 80 $\times$ 80 from 200 subjects (about 7 images each subject). The images of each individuals were taken under different expression and illumination conditions, marked by `ba', `bd', `be', `bf', `bg', `bj', and `bk'.
To represent a facial image, similar to the work \citep{YangZhangShiuZhang2013}, we created a 43$\times$43 region covariance matrix, i.e., a specific SPD matrix, which is summarised over intensity value, spatial coordinates, 40 Gabor filters at 8 orientations and 5 scales.

We first tested KSSCR on seven subsets from FERET database which randomly covers some different clusters. The clustering results of the tested approaches are shown in Figure \ref{fig3} and \ref{fig4} with varying number of clusters. From Figure \ref{fig3}, we observe that the clustering accuracy of the proposed approach is better than other state-of-the-art methods in most cases, peaking on 30-cluster subset with $\lambda = 0.1$, $\rho = 3.0$, and $\gamma= 2 \times 10^{-3}$. On average, the clustering rate on all seven subsets are 44.49, 49.33, 78.56, 47.54, 72.50, and 80.42 for K-means, SSC, LRSC, LRR, KSSCE, and KSSCR, respectively. In Figure \ref{fig4}, the NMI is presented to show the performance of different methods. The average scores are 66.65, 68.31, 88.43, 69.08, 83.01, and 88.46 for K-means, SSC, LRSC, LRR, KSSCE, and KSSCR, respectively. As can be seen, KSSCR achieves the comparable results and a little bit better than that of LRSC. While compared to KSSCE, KSSCR is leading by a large margin. This further verifies the advantage of fully considering the intrinsic geometry structure within data.
\begin{figure}[ht]
\centering
  \includegraphics[width=0.5 \textwidth]{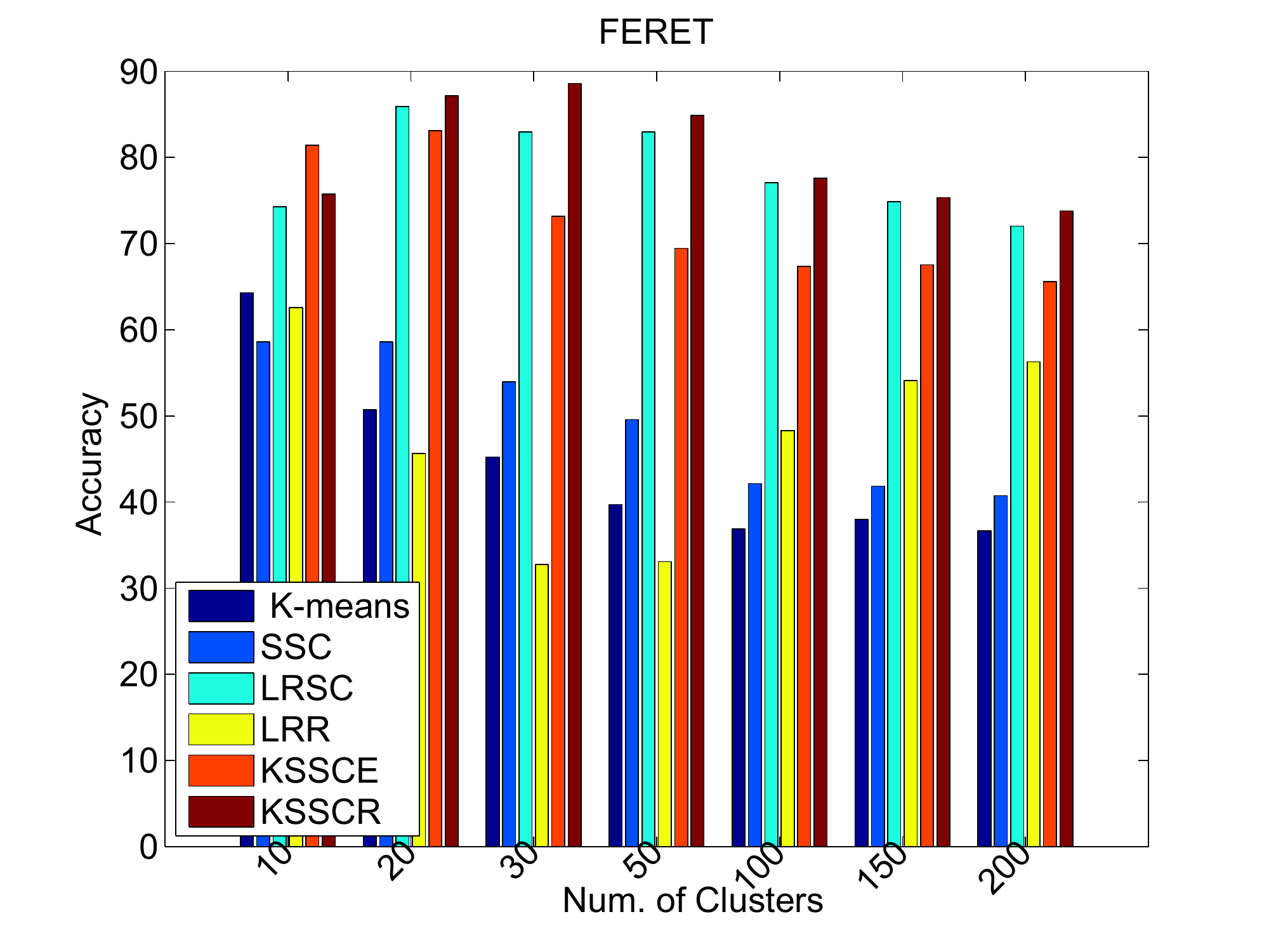}
   \caption{Comparison of clustering accuracy (\%) on FERET dataset. The average scores on all 7 tests are
   44.49,  49.33, 78.56,  47.54, 72.50, and  80.42 for K-means, SSC, LRSC, LRR, KSSCE, and KSSCR, respectively. }\label{fig3}
\end{figure}
\begin{figure}[ht]
\centering
 \includegraphics[width=0.5 \textwidth]{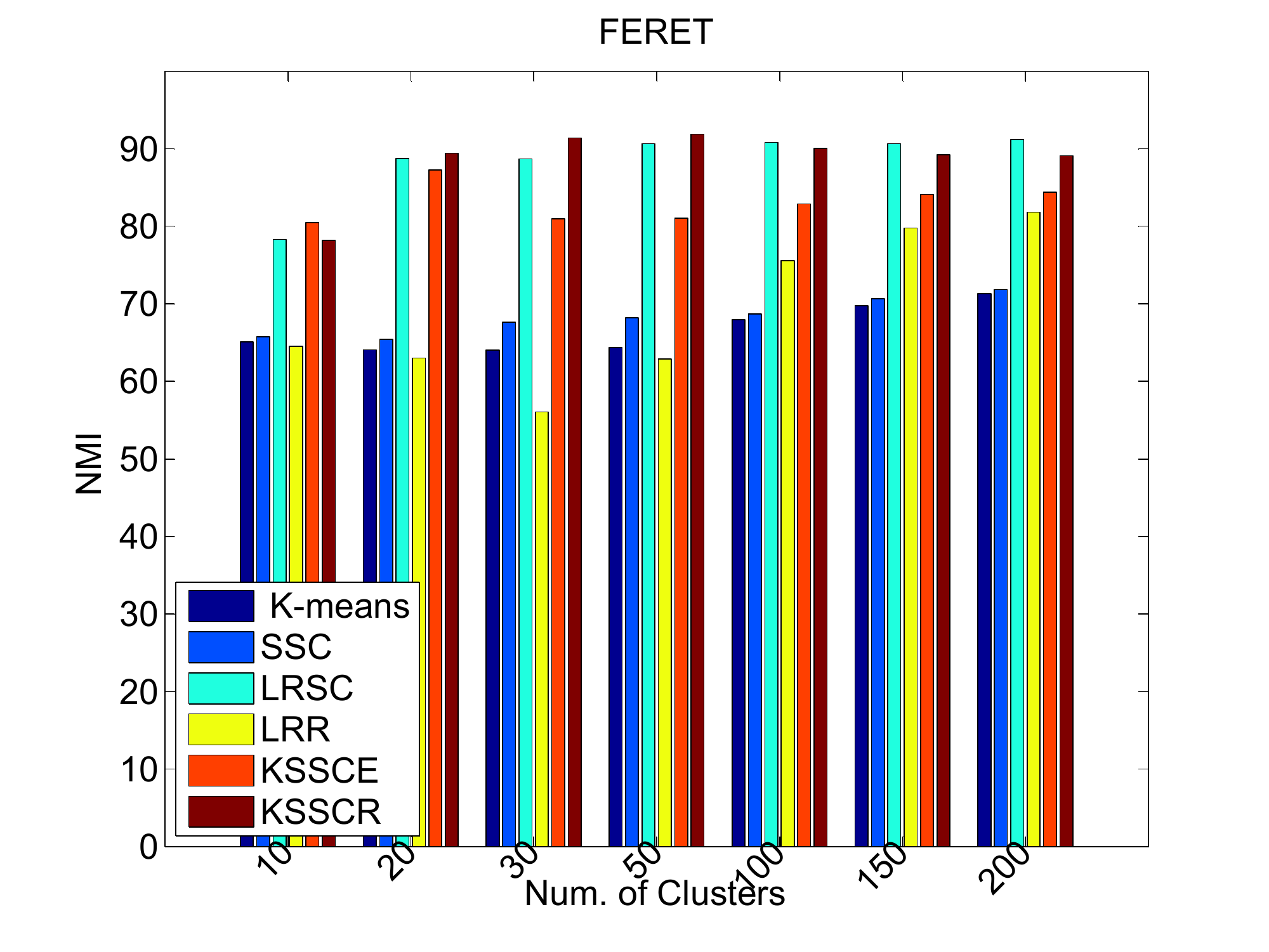}
   \caption{Comparison of NMI  (\%) on FERET dataset. The average scores on all 7 tests are 66.65, 68.31, 88.43, 69.08, 83.01, and  88.46 for K-means, SSC, LRSC, LRR, KSSCE, and KSSCR, respectively. }\label{fig4}
\end{figure}

Next, we tested the effect of parameter $\gamma$ in Log-Euclidean Gaussian kernel by fixing the number of clusters to 30. Fig. \ref{fig5} shows the clustering performance versus parameter $\gamma$ on the FERET dataset. From the figure, we can see that the clustering score increases as $\gamma$ gets larger, reaching peak at about $2\times10^{-3}$ and decreasing afterwards. This helps to determine the
value of parameter $\gamma$ in the clustering experiments.
\begin{figure}[ht]
\centering
  \includegraphics[width=0.45 \textwidth]{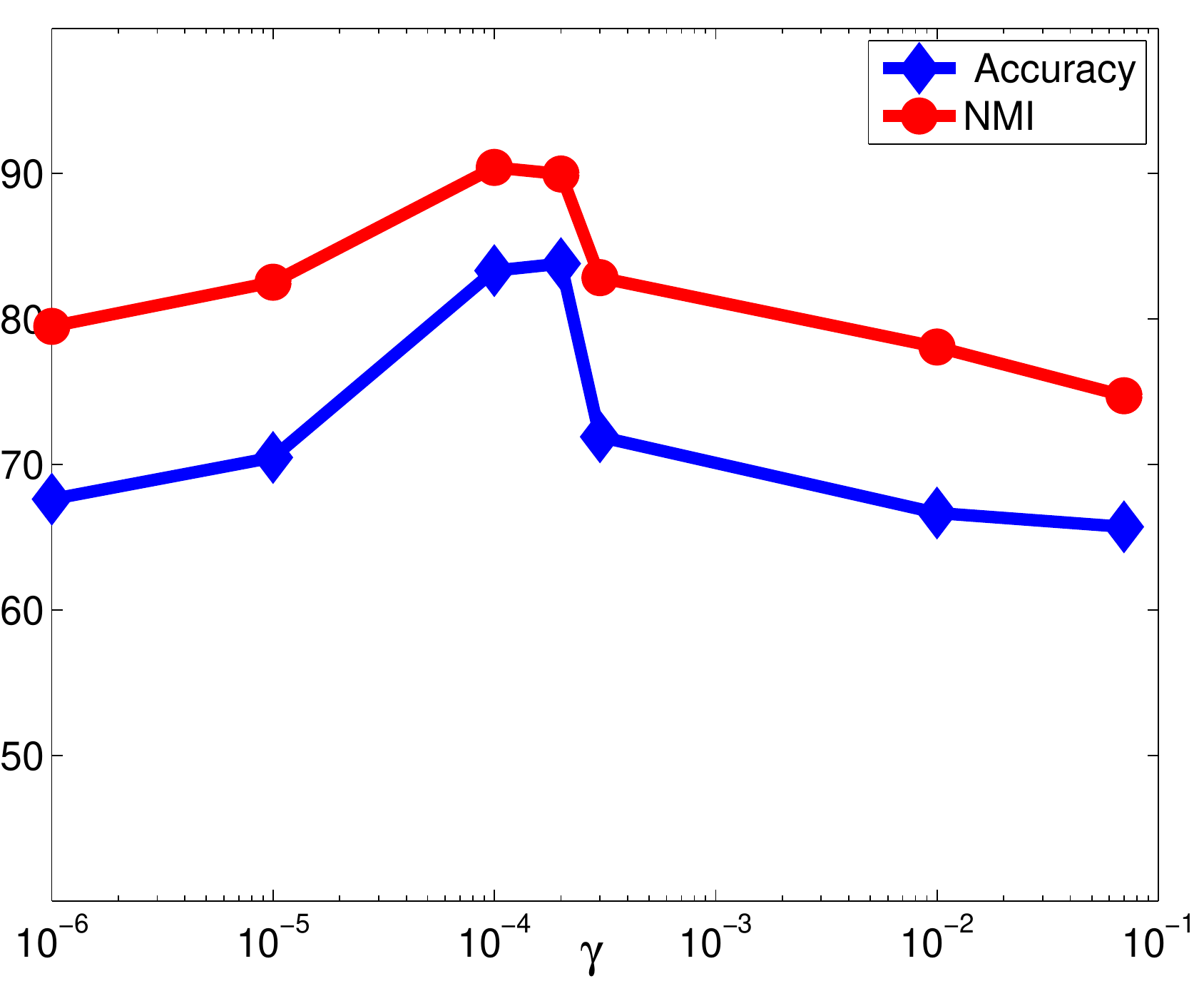} 
   \caption{Accuracy and NMI (\%) (y-axis) of KSSCR with different $\gamma$ (\text{x}-axis) on FERET dataset. }\label{fig5}
\end{figure}

\section{Conclusion}\label{SectionV}
In this paper, we proposed a novel algorithm called kernel sparse subspace clustering (KSSCR) for SPD matrices, a special type of data that cripples original sparse representation methods based on linear reconstruction. By using an appropriate Log-Euclidean kernel, we exploited the {\it{self-expressive}} property to obtain  sparse representation of the original data mapped into kernel reproducing Hilbert space based on Riemannian metric. Experimental results show that the proposed method can provide better clustering solutions than the state-of-the-art approaches thanks to incorporating Riemannian geometry structure.

\section*{Acknowledgement}
The funding information is hid for the purpose of the anonymous review process.


\begin{thebibliography}{10}\itemsep=-1pt

\bibitem{ArsignyFillardPennecAyache2007}
V.~Arsigny, P.~Fillard, X.~Pennec, and N.~Ayache.
\newblock Geometric means in a novel vector space structure on symmetric
  positive-definite matrices.
\newblock {\em {SIAM} Journal on Matrix Analysis and Applications},
  29(1):328--347, 2007.

\bibitem{BoydParikhChuPeleatoEckstein2011}
S.~Boyd, N.~Parikh, E.~Chu, B.~Peleato, and J.~Eckstein.
\newblock {\em Distributed Optimization and Statistical Learning via the
  Alternating Direction Method of Multipliers}, volume~3.
\newblock Foundations and Trends in Machine Learning, 2011.

\bibitem{CherianSra2014}
A.~Cherian and S.~Sra.
\newblock Riemannian sparse coding for positive definite matrices.
\newblock In {\em Proceedings of ECCV}, volume 8691, pages 299--314. Springer
  International Publishing, 2014.

\bibitem{CherianSraBanerjeePapanikolopoulos2013}
A.~Cherian, S.~Sra, A.~Banerjee, and N.~Papanikolopoulos.
\newblock Jensen-bregman logdet divergence with application to efficient
  similarity search for covariance matrices.
\newblock {\em IEEE Transactions on Pattern Analysis and Machine Intelligence},
  35(9):2161--2174, Sept 2013.

\bibitem{DonohoEladTemlyakov2006}
D.~Donoho, M.~Elad, and V.~Temlyakov.
\newblock Stable recovery of sparse overcomplete representations in the
  presence of noise.
\newblock {\em IEEE Transactions on Information Theory}, 52(1):6--18, 2006.

\bibitem{ElhamifarVidal2013}
E.~Elhamifar and R.~Vidal.
\newblock Sparse subspace clustering: Algorithm, theory, and applications.
\newblock {\em IEEE Transactions on Pattern Analysis and Machine Intelligence},
  35(11):2765--2781, 2013.

\bibitem{FuGaoHongTien2015}
Y.~Fu, J.~Gao, X.~Hong, and D.~Tien.
\newblock Low rank representation on {R}iemannian manifold of symmetric
  positive deffinite matrices.
\newblock In {\em Proceedings of SDM}, 2015, DOI:10.1137/1.9781611974010.36.

\bibitem{HarandiSandersonHartleyLovell2012}
M.~Harandi, C.~Sanderson, R.~Hartley, and B.~Lovell.
\newblock Sparse coding and dictionary learning for symmetric positive definite
  matrices: A kernel approach.
\newblock In {\em Proceedings of ECCV}, pages 216--229, 2012.

\bibitem{HarandiHartleyLovellSanderson2014}
M.~T. Harandi, R.~Hartley, B.~C. Lovell, and C.~Sanderson.
\newblock Sparse coding on symmetric positive definite manifolds using bregman
  divergences.
\newblock {\em IEEE Transactions on Neural Networks and Learning Systems},
  2015, DOI:10.1109/TNNLS.2014.2387383.

\bibitem{HoXieVemuri2013}
J.~Ho, Y.~Xie, and B.~C. Vemuri.
\newblock On a nonlinear generalization of sparse coding and dictionary
  learning.
\newblock In {\em Proceedings of ICML}, volume~28, pages 1480--1488, 2013.

\bibitem{JayasumanaHartleySalzmannLiHarandi2013}
S.~Jayasumana, R.~Hartley, M.~Salzmann, H.~Li, and M.~T. Harandi.
\newblock Kernel methods on the {R}iemannian manifold of symmetric positive
  definite matrices.
\newblock In {\em Proceedings of CVPR}, pages 73--80, June 2013.

\bibitem{LiWangZuoZhang2013}
P.~Li, Q.~Wang, W.~Zuo, and L.~Zhang.
\newblock Log-{E}uclidean kernels for sparse representation and dictionary
  learning.
\newblock In {\em Proceedings of ICCV}, pages 1601--1608, Dec 2013.

\bibitem{LinLiuLi2015}
Z.~Lin, R.~Liu, and H.~Li.
\newblock Linearized alternating direction method with parallel splitting and
  adaptive penalty for separable convex programs in machine learning.
\newblock {\em Machine Learning}, 99(2):287--325, 2015.

\bibitem{LiuLinYanSunMa2013}
G.~Liu, Z.~Lin, S.~Yan, J.~Sun, and Y.~Ma.
\newblock Robust recovery of subspace structures by low-rank representation.
\newblock {\em IEEE Transactions on Pattern Analysis and Machince
  Intelligence}, 35(1):171 -- 184, Jan. 2013.

\bibitem{LiuLinYu2010}
G.~Liu, Z.~Lin, and Y.~Yu.
\newblock Robust subspace segmentation by low-rank representation.
\newblock In {\em Proceedings of ICML}, pages 663--670, 2010.

\bibitem{NguyenYangShenSun2015}
H.~Nguyen, W.~Yang, F.~Shen, and C.~Sun.
\newblock Kernel low-rank representation for face recognition.
\newblock {\em Neurocomputing}, 155:32--42, 2015.

\bibitem{PatelVidal2014}
V.~M. Patel and R.~Vidal.
\newblock Kernel sparse subspace clustering.
\newblock In {\em Proceedings of ICIP}, pages 2849--2853, Oct 2014.

\bibitem{PennecFillardAyache2006}
X.~Pennec, P.~Fillard, and N.~Ayache.
\newblock A {R}iemannian framework for tensor computing.
\newblock {\em International Journal Of Computer Vision}, 66:41--66, 2006.

\bibitem{SivalingamBoleyMorellasPapanikolopoulos2014}
R.~Sivalingam, D.~Boley, V.~Morellas, and N.~Papanikolopoulos.
\newblock Tensor sparse coding for positive definite matrices.
\newblock {\em IEEE Transactions on Pattern Analysis and Machine Intelligence},
  36(3):592--605, 2014.

\bibitem{Sra2012}
S.~Sra.
\newblock A new metric on the manifold of kernel matrices with application to
  matrix geometric means.
\newblock In F.~Pereira, C.~Burges, L.~Bottou, and K.~Weinberger, editors, {\em
  Proceedings of NIPS}, pages 144--152, 2012.

\bibitem{TuzelPorikliMeer2006}
O.~Tuzel, F.~Porikli, and P.~Meer.
\newblock {Region Covariance: A Fast Descriptor for Detection And
  Classification}.
\newblock In {\em Proceedings of ECCV}, pages 589--600, 2006.

\bibitem{TuzelPorikliMeer2008}
O.~Tuzel, F.~Porikli, and P.~Meer.
\newblock Pedestrian detection via classification on {R}iemannian manifolds.
\newblock {\em IEEE Transactions on Pattern Analysis and Machine Intelligence},
  30(10):1713--1727, Oct 2008.

\bibitem{Vidal2011}
R.~Vidal.
\newblock Subspace clustering.
\newblock {\em IEEE Signal Processing Magazine}, 28(2):52--68, 2011.

\bibitem{VidalFavaro2014}
R.~Vidal and P.~Favaro.
\newblock Low rank subspace clustering ({LRSC}).
\newblock {\em Pattern Recognition Letters}, 43(1):47--61, 2014.

\bibitem{WangHuGaoSunYin2015a}
B.~Wang, Y.~Hu, J.~Gao, Y.~Sun, and B.~Yin.
\newblock Kernelized low rank representation on {G}rassmann manifolds.
\newblock {\em arXiv preprint arXiv:1504.01806}.

\bibitem{WangHuGaoSunYin2015}
B.~Wang, Y.~Hu, J.~Gao, Y.~Sun, and B.~Yin.
\newblock Low rank representation on {G}rassmann manifolds: An extrinsic
  perspective.
\newblock {\em arXiv preprint arXiv:1504.01807}.

\bibitem{WangHuGaoSunYin2014}
B.~Wang, Y.~Hu, J.~Gao, Y.~Sun, and B.~Yin.
\newblock Low rank representation on {G}rassmann manifold.
\newblock In {\em Proceedings of ACCV}, pages 81--96, Singapore, November 2014.

\bibitem{YangZhangShiuZhang2013}
M.~Yang, L.~Zhang, S.~C. Shiu, and D.~Zhang.
\newblock Gabor feature based robust representation and classification for face
  recognition with {G}abor occlusion dictionary.
\newblock {\em Pattern Recognition}, 46(7):1865 -- 1878, 2013.

\bibitem{YinGaoGuo2015}
M.~Yin, J.~Gao, and Y.~Guo.
\newblock Nonlinear low-rank representation on {S}tiefel manifolds.
\newblock {\em Electronics Letters}, 51(10):749--751, 2015.

\end{thebibliography}

\end{document}